\documentclass[sigconf]{acmart}
\usepackage{amsmath,amsfonts}
\usepackage{multirow}
\usepackage{makecell}
\usepackage{balance}
\AtBeginDocument{%
  }

\begin{document}

\title{MambaHash: Visual State Space Deep Hashing Model for Large-Scale Image Retrieval}

\author{Chao He}
\email{3089840275@qq.com}
\affiliation{%
	\institution{School of Computer Science, Inner Mongolia University, Provincial Key Laboratory of Mongolian Information Processing Technology, National and Local Joint Engineering Research Center of Mongolian Information Processing}
	\streetaddress{No.235 West College Road,Saihan Distric}
	\city{Hohhot}
	\country{China}
}

\author{Hongxi Wei}
\authornotemark[1]
\email{cswhx@imu.edu.cn}
\affiliation{%
	\institution{School of Computer Science, Inner Mongolia University, Provincial Key Laboratory of Mongolian Information Processing Technology, National and Local Joint Engineering Research Center of Mongolian Information Processing}
	\streetaddress{No.235 West College Road,Saihan Distric}
	\city{Hohhot}
	\country{China}
}

\begin{abstract}
  Deep image hashing aims to enable effective large-scale image retrieval by mapping the input images into simple binary hash codes through deep neural networks. More recently, Vision Mamba with linear time complexity has attracted extensive attention from researchers by achieving outstanding performance on various computer tasks. Nevertheless, the suitability of Mamba for large-scale image retrieval tasks still needs to be explored. Towards this end, we propose a visual state space hashing model, called MambaHash. Concretely, we propose a backbone network with stage-wise architecture, in which grouped Mamba operation is introduced to model local and global information by utilizing Mamba to perform multi-directional scanning along different groups of the channel. Subsequently, the proposed channel interaction attention module is used to enhance information communication across channels. Finally, we meticulously design an adaptive feature enhancement module to increase feature diversity and enhance the visual representation capability of the model. We have conducted comprehensive experiments on three widely used datasets: CIFAR-10, NUS-WIDE and IMAGENET. The experimental results demonstrate that compared with the state-of-the-art deep hashing methods, our proposed MambaHash has well efficiency and superior performance to effectively accomplish large-scale image retrieval tasks. Source code is available https://github.com/shuaichaochao/MambaHash.git
\end{abstract}


\maketitle

\section{Introduction}
In the last decade, with the rapid development of the Internet and the popularization of mobile media devices, the massive end-users have generated large-scale visual data. Therefore, it is a challenging task to accurately and efficiently retrieve the information people desire from the vast amount of visual data. Of these, large-scale image retrieval has become one of the research hotspots in the field of information retrieval and has attracted extensive attention \cite{zhu2016deep,cui2019scalable,wei2019saliency,brogan2021fast}. Large-scale image retrieval tasks support the retrieval of relevant images from large image databases and are widely applied in scenarios such as search engines, recommender systems, and shopping software, etc. Among all the methods proposed for large-scale image retrieval tasks, hashing is one of the most effective image retrieval methods with its extremely fast speed and low memory usage \cite{zhang2010self, liu2012supervised}. It aims to learn a hash function that maps the image in the high-dimensional pixel space to the low-dimensional Hamming space, while the similarity of the images in the original space can be preserved \cite{guo2017learning}.

Hashing methods generally consist of two phases. In the first phase, it aims to efficiently extract the image features. The second phase utilizes various nonlinear functions to squeeze the image features into binary codes and designs diverse loss functions \cite{li2015feature,fan2020deep,yuan2020central,cao2017hashnet} to guarantee the semantic similarity of the raw image pairs. On the basis of the way they extract image features, existing hashing methods can be summarized into two categories: hand-crafted based methods and deep learning based methods. Hand-crafted based methods \cite{charikar2002similarity,weiss2008spectral} utilize hand-crafted visual descriptors \cite{oliva2001modeling} (i.e., image features) to learn hash function. Nevertheless, hand-crafted features can not guarantee the semantic similarity of the raw image pairs, with resultant performance degradation in the subsequent hash function learning process. In contrast to hand-crafted based methods, deep learning based methods \cite{lin2015deep,zheng2020deep,zhang2019improved,li2015feature} utilize  deep convolutional neural networks (CNNs) (e.g. AlexNet, ResNet) to efficiently extract image features in an end-to-end fashion, which have become the most dominant approach in the field of image retrieval.

\begin{figure}[t]
	\centering
	\includegraphics[width=0.75\linewidth,height=0.8\linewidth]{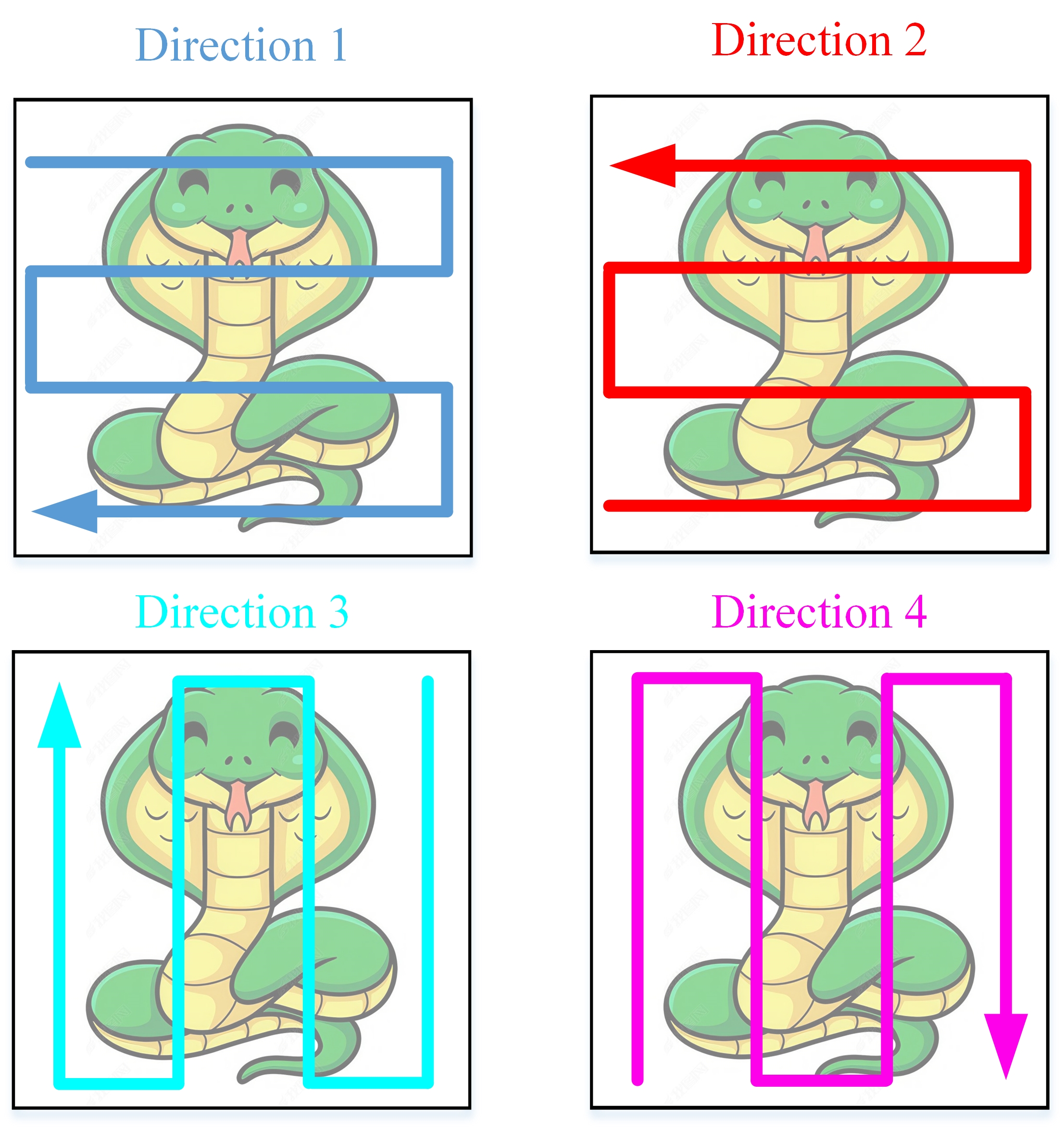}
	\caption{The direction of scanning along the four groups of channels with Mamba is illustrated, which enables a greater focus on global contextual information.}
\end{figure}

In recent years, motivated by the latest advancement in Vision Transformer (ViT) \cite{dosovitskiy2020image}, many ViT-based hashing methods \cite{chen2022transhash,li2022hashformer,li2023msvit,xiao2024deep} have been proposed. However, the original ViT excels at capturing long-range dependencies, which tends to ignore local features. Moreover, the self-attention of ViT is related quadratically to the sequence length, which brings significant computational challenges. In comparison to ViT, CNNs have a more robust ability to extract local features. It has recently been demonstrated that it is beneficial to combine the multi-head self-attention in ViT with the convolutional layer in CNNs \cite{park2022vision}. Therefore, many hybrid hashing networks have been proposed \cite{he2024hybridhash,zhu2023multi} by taking advantage of ViT to capture long-range dependencies, and of CNN to extract local information. Hybrid hashing networks have achieved significant performance improvement, yet the quadratic complexity problem that is inherent to self-attention in ViT still exists. Recently, many works \cite{chu2021twins,tu2022maxvit,shaker2023swiftformer,gu2023mamba} have been devoted to mitigating the quadratic complexity inherent in self-attention. Of these, State Space Models (SSMs) \cite{gu2021combining} has attracted extensive attention from researchers which capture the intricate dynamics and inter-dependencies within linguistic sequences \cite{gu2021efficiently}. One notable method in this area is the structured state-space sequence model (S4) \cite{gu2021efficiently}, which aims to address long-distance dependencies while maintaining linear complexity. More recently, Mamba \cite{gu2023mamba} introduces an input-dependent SSM layer and leverages the parallel selective scanning mechanism (S6) to model long-range dependencies with linear complexity, which is widely regarded as the toughest competitor to Transformer \cite{abdelrahman2024groupmamba,liu2024vmamba}. In the computer vision domain, many works have been presented applying Mamba to various tasks, such as image classification \cite{abdelrahman2024groupmamba,liu2024vmamba,zhu2024vision}, object detection \cite{behrouz2024mambamixer,yang2024plainmamba}, image segmentation \cite{liu2024swin,ma2024u} and so on. It demonstrates the effectiveness of Mamba in the area of computer vision. Some works \cite{yu2024mambaout} have further investigated the nature of Mamba and analyzed that Mamba is ideally suited for tasks with two key characteristics: long sequences and autoregression. Unfortunately, image retrieval tasks do not possess the above both characteristics. Therefore, it is worth exploring whether Mamba can be applied into image retrieval tasks.

In this paper, we argue that Mamba is still available for image retrieval tasks with well designed network architecture. To this end, we propose a visual state space deep hashing model called \textbf{MambaHash}. Specifically, with regard to pairwise hash learning, we design a Mamba-based backbone network. The design idea of modulated group Mamba layers \cite{abdelrahman2024groupmamba} is introduced, where the number of input channels is divided into four groups, and Mamba is utilized to perform multi-direction scanning (\textbf{As shown in Figure 1}) along the four groups of channels for effective modeling of local and global information, thereby enhancing the computational efficiency. Since the grouping operation limits the information interaction between channels, we propose a Channel Interaction Attention Module (CIAM) to enhance cross channel communication. Moreover, we adopt the stage-wise architecture similar to CNNs \cite{guo2022cmt} for gradually decreasing the resolution, and elaborately design an Adaptive Feature Enhancement Module (AFEM) to enhance the diversity of visual features. Finally, average pooling is utilized to replace class tokens in MambaHash for obtaining image features, followed by hash layer to output binary codes. To preserve the semantic similarity of image pairs in feature space, we adopt maximum likelihood estimation to pull close similar pairs and push away dissimilar pairs in Hamming space. To control the quantization error for transforming continuous embedded features into binary codes, we further introduce pairwise quantization loss \cite{zhu2016deep}. 

In general, the main contributions of this paper are summarized as follows:
\begin{itemize}
	\item A novel visual state space deep hashing model (MambaHash) is proposed, which utilizes Mamba for multi-direction scanning through channel grouping operations to efficiently model local and global information with linear time complexity.
	\item To enhance cross-channel communication for improved feature aggregation, a channel interaction attention module (CIAM) is proposed.
	\item An adaptive feature enhancement module (AFEM) is elaborately designed to increase the diversity of visual features and enhance the visual representation.
	\item We perform comprehensive experiments on three widely-studied datasets (CIFAR-10 \cite{krizhevsky2009learning}, NUS-WIDE \cite{chua2009nus}, and IMAGENET \cite{russakovsky2015imagenet}). Experimental results indicate that our proposed MambaHash has superior performance compared to state-of-the-art deep supervised hashing methods.
\end{itemize}

\section{Related Works}
\begin{figure*}[ht]
	\centering
	\includegraphics[width=0.9\linewidth]{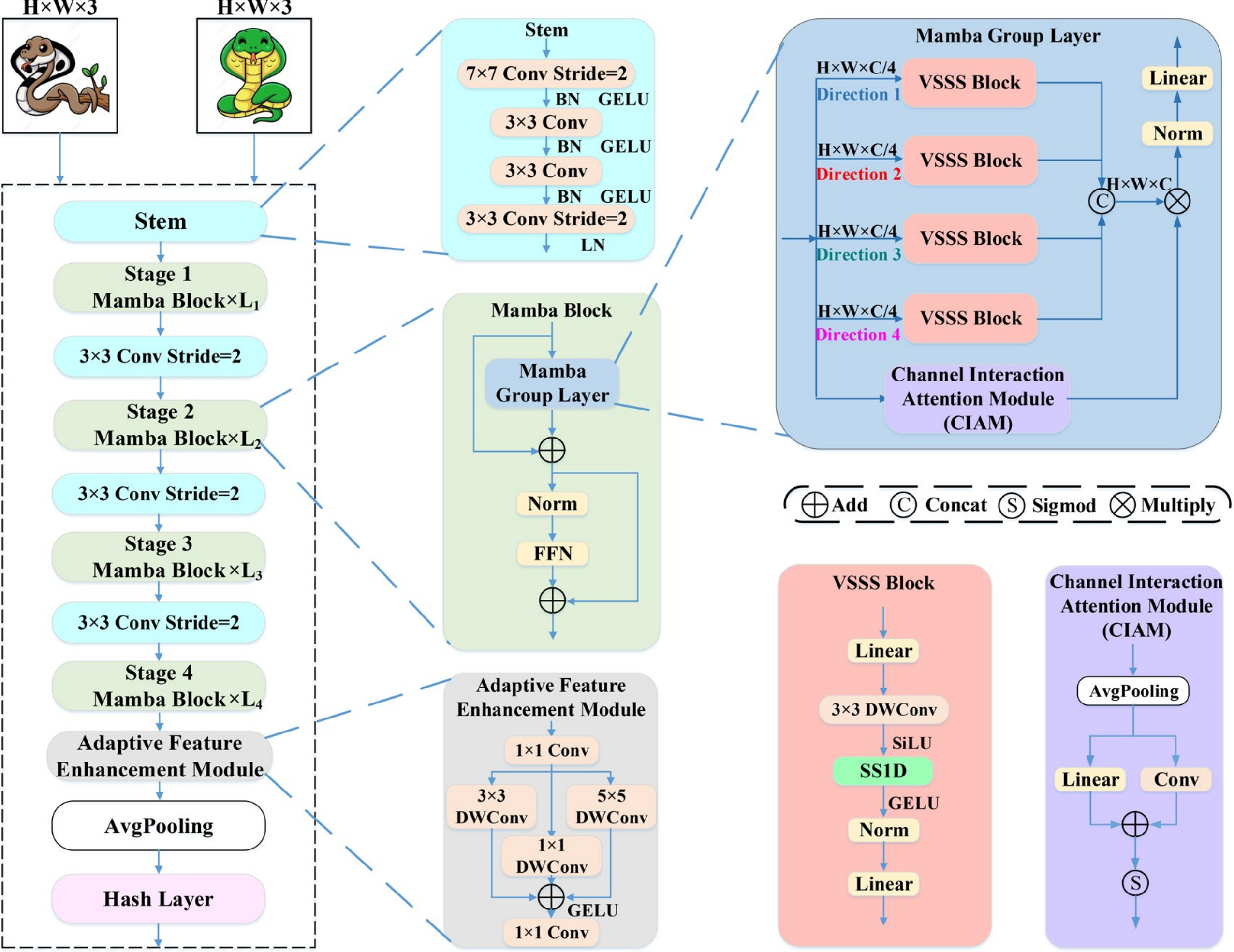}
	\caption{The detailed architecture of the proposed MambaHash. MambaHash accepts pairwise images as input, and adopts a similar stem architecture \cite{he2019bag} to divide the images into overlapping patches with the generated patches fed into the Mamba block. The whole model architecture consists of four stages, followed by an Adaptive feature enhancement module to increase feature diversity. Finally, the binary codes are output after the hashing layer.}
\end{figure*}

\subsection{General Overview of Vision Mamba}
Since self-attention in Vision Transformer (ViT) \cite{dosovitskiy2020image} has a quadratic time complexity, many researchers have devoted efforts to mitigate the inherent quadratic complexity of self-attention. Among them, state space models (SSMs) have attracted extensive attention. S4ND \cite{nguyen2022s4nd} is pioneer study to process visual data as continuous signals in 1D, 2D and 3D domains by employing SSM blocks. Recently, motivated by the development of Mamba \cite{gu2023mamba}, many Mamba-based works have been proposed in the visual domain. Vision Mamba (Vim) \cite{zhu2024vision} adopts the similar design as ViT and utilizes Mamba to perform a single bidirectional scan of the image. During the same period, VMamba \cite{liu2024vmamba} utilizes Mamba to perform a bidirectional scan of the image across two spatial dimensions in a stage-wise style design \cite{guo2022cmt}, effectively building a global receptive field. Subsequently, LocalMamba \cite{huang2024localmamba} divides the image into distinct windows, in which Mamba is utilized to perform a single directional scan of the image, effectively capturing local dependencies while maintaining a global perspective. PlainMamba \cite{yang2024plainmamba} proposes a continuous 2D scanning approach and encodes the directionality of each scanning order to improve spatial and semantic continuity. The above methods overlook information interaction across channels, which limits the ability of Mamba to model global information in multi-dimensional data. MambaMixer \cite{behrouz2024mambamixer} proposes a dual selection mechanism across tokens and channels, which demonstrates the importance of mixing channels and further improves performance. GroupMamba \cite{abdelrahman2024groupmamba} designs a modulated group Mamba layer, which achieves comprehensive spatial coverage and effective local and global information modeling by using multi-direction scanning methods and introduces channel affinity modulation attention to enhance the ability to communicate across channels. These models have been applied for image classification, image segmentation \cite{liu2024swin,ma2024u} and various other tasks \cite{he2025pan,wang2024graph}. Their wide applicability demonstrates the effectiveness of Mamba in the visual domain.

\subsection{Deep Supervised Hashing for Image Retrieval}
Image hashing maps images into simple binary codes. Early works \cite{lin2014fast,zhang2010self} on image retrieval introduced hashing to effectively address the problems of high storage usage and low retrieval speed. With the development of CNN technology, many deep supervised hashing methods based on CNN have been proposed. DPSH \cite{li2015feature} is the first deep hashing method that can utilize pairwise images for simultaneous feature learning and hash code learning. HashNet \cite{cao2017hashnet} performs hashing via a continuous method with convergence guarantees and proposes a weighted pairwise cross-entropy loss function. It addresses the problem of ill-posed gradients and data imbalance in an end-to-end framework for deep feature learning and binary hash coding. Subsequently, DCH \cite{cao2018deep} proposes a pairwise cross-entropy loss based on the Cauchy distribution, which significantly penalizes similar image pairs with Hamming distance larger than a given Hamming radius threshold. MMHH \cite{kang2019maximum} proposes a max-marginal t-distribution loss, where the t-distribution concentrates more similiar points within the Hamming ball, and the margins characterize the Hamming radius, resulting in less penalization of similiar points that fall inside the Hamming ball. DPN \cite{fan2020deep} further proposes a new polarization loss. It has guaranteed to minimize the original Hamming distance-based loss without quantization error, while avoiding complex binary optimization solving. 

Inspired by the recent advancement of ViT, Transhash \cite{chen2022transhash} proposes a purely Transformer-based deep hash learning framework and innovates dual-stream multi-granular feature learning on top of the Transformer to learn discriminative global and local features. HashFormer \cite{li2022hashformer} further utilizes ViT as a backbone network and regards binary codes as intermediate representations of the proxy task and proposes average precision loss. Shortly after, MSViT \cite{li2023msvit} utilizes ViT to process image patches with different granularity to obtain features at different scales and fuse them efficiently. HybridHash \cite{he2024hybridhash} proposes a new deep hash hybrid network that takes advantage of Transformers to capture long-range dependencies and of CNNs to extract local information. Nevertheless, the quadratic complexity problem inherent in self-attention from ViT still exists. Therefore, Mamba-based deep hashing methods are worth being explored.

\section{MambaHash}
\subsection{Preliminaries}
SSMs are termed as linear time-invariant systems, which map an input sequence $ x\left(t\right) \in \mathbb{R}^L $ to a response sequence $ y\left(t\right) \in \mathbb{R}^L $ via the hidden state $ h\left(t\right) \in \mathbb{R}^N$ \cite{gu2023mamba}. More specifically, continuous-time SSMs can be formulated as linear ordinary differential equations (ODEs) as follows:
\begin{equation}
	\begin{aligned}
		h^{\prime}(t) & =\mathbf{A} h(t)+\mathbf{B} x(t) \\
		y(t) & =\mathbf{C} h(t)
	\end{aligned}
\end{equation}
Where $ h\left(t\right) $ is the current hidden state, $ h^{\prime}(t) $ is the updated hidden state, $ x\left(t\right) $ is the current input, $ y\left(t\right) $ is the output, $ \mathbf{A} \in \mathbb{R}^{N \times N}$ is SSM’s evolution matrix, and $ \mathbf{B} \in \mathbb{R}^{N \times 1} $, $ \mathbf{C} \in \mathbb{R}^{N \times 1} $ are the input and output projection matrices, respectively. 

To be integrated into deep models, the SSMs need to be converted from a continuous-time function-to-function map to a discrete-time sequence-to-sequence map. Therefore, the discrete state space models S4 \cite{gu2021efficiently} and Mamba \cite{gu2023mamba} utilize the parameter $ \boldsymbol{\Delta} $ to discretize the above system. This discretization is typically
done through the Zero-Order Hold (ZOH) method given in Equation 2.
\begin{equation}
	\begin{array}{l}
		h_{t}=\overline{\mathbf{A}} h_{t-1}+\overline{\mathbf{B}} x_{t}, \\
		y_{t}=\mathbf{C} h_{t},
	\end{array}
\end{equation}
Where
\begin{equation}
	\begin{array}{l}
		\overline{\mathbf{A}}=\exp (\boldsymbol{\Delta} \mathbf{A}) \\
		\overline{\mathbf{B}}=(\boldsymbol{\Delta} \mathbf{A})^{-1}(\exp (\boldsymbol{\Delta} \mathbf{A}-I)) \cdot \boldsymbol{\Delta} \mathbf{B}
	\end{array}
\end{equation}
While S4 \cite{gu2021efficiently} and Mamba \cite{gu2023mamba} utilize similar discretization methods, Mamba proposes an S6 selective scanning mechanism that is distinguished from S4 by conditioning parameters $ \boldsymbol{\Delta} \in \mathbb{R}^{B \times L \times D}$, $ \mathbf{B} \in \mathbb{R}^{B \times L \times N} $, and $ \mathbf{C} \in \mathbb{R}^{B \times L \times N} $, on the input $ x \in \mathbb{R}^{B \times L \times D}$, where $ B $ is the batch size, $ L $ is the sequence length, and $ D $ is the feature dimension. 

\subsection{Overall Architecture}
The overall architecture of MambaHash is represented in Figure 2, which accepts pairwise input images $ \left\{\boldsymbol{x}_i, \boldsymbol{x}_j \right\} $ and represents the images as binary codes. MambaHash mainly integrates three key components, which are Mamba Block, Adaptive Feature Enhancement Module, and Hash Layer.

Given an image $ \boldsymbol{x}_i \in \mathbb{R}^{H\times W\times 3} $, as opposed to the previous non-overlapping patch way of ViT, we adopt a $ 7 \times 7 $ convolution (with a stride of 2) and a stem structure \cite{guo2022cmt} of 64 output channels to reduce the resolution of the image, followed by two $ 3 \times 3 $ convolutions, finally using a $ 3 \times 3 $ convolution (with a stride of 2) in an overlapping way to better extract the local information. Subsequently, the obtained image features maps $ \boldsymbol{x}_o \in \mathbb{R}^{H/4\times W/4\times D} $ are reshaped into $ \boldsymbol{x}_o \in \mathbb{R}^{N \times D} $ and go through four stages of the Mamba block for hierarchical feature representation extraction, where $ N=H/4\times W/4 $ is the sequence length of the Mamba Block input and $ D $ is the number of channels. The adaptive feature enhancement module extracts multi-scale image features with convolutional kernels of different sizes and increases feature diversity. Finally hash codes are output through hash layer.

\subsection{Mamba Block}
As the core component of MambaHash, Mamba block is utilized to extract global context information and local information. The Mamba Block consists of Mamba Group Layer, and a Feed-Forward Network (FFN) with skip connections \cite{he2016deep} and LayerNorm (LN) \cite{ba2016layer}. The detailed process can be described as follows:
\begin{equation}
	\begin{array}{l}
		\boldsymbol{x}_{t} =\boldsymbol{x}_o + MambaGroupLayer\left(\boldsymbol{x}_o \right)\\
		\boldsymbol{x}_{mg} = \boldsymbol{x}_{t}+FFN\left(LN\left(\boldsymbol{x}_{t}\right)\right)
	\end{array}
\end{equation}
where $ \boldsymbol{x}_{t} $ denotes the mid feature output and $ \boldsymbol{x}_{mg} $ denotes the output of the Mamba block.

\subsubsection{Mamba Group Layer}
The Mamba group layer mainly contains the VSSS Block \cite{abdelrahman2024groupmamba} and the Channel Interaction Attention Module (CIAM). Moreover, since Mamba is computationally inefficient on large number of channels $ D $ in the input sequence, we have introduced a grouped Mamba operation \cite{abdelrahman2024groupmamba} in the Mamba groups layer. Specifically, the input channels are divided into 4 groups, the size of each group is $ D/4 $, and a separate VSSS block is applied to each group. For the purpose of spatial dependency modeling, we utilize the VSSS block to perform the scanning of the four groups in each of the four directions, which can better capture the global context information. The four directions are: left-to-right, right-to-left, top-to-bottom, and bottom-to-top, as shown in Figure 1. We have formed four sequences from the input $ \boldsymbol{x}_o $, denoted $ \boldsymbol{x}_L $, $ \boldsymbol{x}_R $, $ \boldsymbol{x}_T $, $ \boldsymbol{x}_B $, representing one of the four directions specified earlier. Then these sequences are flattened into a single sequence of shape $ \left(N \times D/4\right) $, which is further fed into the VSSS block.
\begin{equation}
	\begin{aligned}
		\boldsymbol{x}_{vs} = Concat( & VSSSBlock(\boldsymbol{x}_L), VSSSBlock(\boldsymbol{x}_R), \\
		& VSSSBlock(\boldsymbol{x}_T), VSSSBlock(\boldsymbol{x}_B)) \\	
	\end{aligned}
\end{equation}

The channel grouping is performed, leading to lack of information interaction between channels. Therefore, CIAM is utilized for cross-channel information communication to enhance the representation capability of the network.
\begin{equation}
	\begin{aligned}
		\boldsymbol{x}_{mgl} = LN(\boldsymbol{x}_{vs} \cdot CIAM(\boldsymbol{x}_o))\boldsymbol{W}_1+\boldsymbol{b}_1	
	\end{aligned}
\end{equation}
where $ \boldsymbol{W}_1 $, and $ \boldsymbol{b}_1 $ are weight and bias for the linear projection.
\subsubsection{VSSS Block} As shown in the pink box in Figure 2, the VSSS block consists of mainly two core components, SS1D \cite{liu2024vmamba} and Depth-Wise Convolution. SS1D is used to globally model the sequence yet neglects the local relationships in the sequence. To alleviate the limitation, we utilize depth-wise convolution to extract local information. The above process can be defined as:
\begin{equation}
	\begin{aligned}
		\boldsymbol{x}_{vsss} = LN(SS1D(DWConv_{3 \times 3}(\boldsymbol{x}_L \boldsymbol{W}_2+\boldsymbol{b}_2)))\boldsymbol{W}_3+\boldsymbol{b}_3			
	\end{aligned}
\end{equation}
where $ \boldsymbol{W}_2 $, $ \boldsymbol{b}_2 $, $ \boldsymbol{W}_3 $ and $ \boldsymbol{b}_3 $ are weights and biases for the linear projections, $ DWConv(*) $ denotes the depth-wise convolution.

\subsubsection{Channel Interaction Attention Module} 
For cross-channel information communication to enhance the feature representation of the network, we propose a channel interaction attention module (CIAM), as illustrated in the purple box in Figure 2. At first, we perform average pooling of channels to calculate channel statistics.
\begin{equation}
	\begin{aligned}
		\boldsymbol{x}_{avg}=AvgPooling(\boldsymbol{x}_o)			
	\end{aligned}
\end{equation}
where $ AvgPooling(*) $ indicates the global average pooling. Next, a two-branch parallelization strategy is adopted, where the first branch is utilized for local information communication between channels. Here we introduce ECA \cite{wang2020eca} to achieve the above objective with a simple fast $ 1D $ convolution of kernel size $ k $.
\begin{equation}
	\begin{aligned}
		\boldsymbol{x}_{local}=Conv1D_{k \times k}(\boldsymbol{x}_{avg})			
	\end{aligned}
\end{equation}
The second branch is used for global information communication between channels. Previous classical channel attention modules \cite{woo2018cbam,hu2018squeeze} use two fully-connected layers to yield inter-channel interactions via channel dimensionality reduction. Nevertheless, avoiding channel dimensionality reduction is important for learning channel attention \cite{wang2020eca}. Therefore, we utilize only one simple fully connected layer for all the interactions between channels.
\begin{equation}
	\begin{aligned}
		\boldsymbol{x}_{global}=\boldsymbol{x}_{avg}\boldsymbol{W}_4+\boldsymbol{b}_4			
	\end{aligned}
\end{equation}
where $ \boldsymbol{W}_4$, and $ \boldsymbol{b}_4 $ are weight and bias for the linear projection. Finally, $ \boldsymbol{x}_{global} $ and $ \boldsymbol{x}_{local} $ are fused and fed into the non-linearity activation function to obtain the importance scores of the channels.
\begin{equation}
	\begin{aligned}
		x_{score}=\sigma(\boldsymbol{x}_{global}+\boldsymbol{x}_{local})			
	\end{aligned}
\end{equation}
where $ \sigma $ represents Sigmod function.

\subsection{Adaptive Feature Enhancement Module}
In this work, we elaborately design an adaptive feature enhancement module (AFEM) to increase feature diversity and enhance the visual representation capability of the model. Moreover, we develop a method for adaptively selecting the channel enhancement ratio to determine the corresponding number of channel increments at different numbers of hash bits. The gray box in Figure 2 illustrates the detailed structure of the adaptive feature enhancement module. Specifically, we first utilize a convolution with the kernel size of 1 to increase the channel dimension according to the channel enhancement ratio. 
\begin{equation}
	\begin{aligned}
		x_{F}=Conv_{1 \times 1}(x_{mgl})			
	\end{aligned}
\end{equation}
Subsequently, convolutions with three different kernel sizes $ 1 \times 1 $, $ 3 \times 3 $, and $ 5 \times 5 $ are utilized to further mine the detailed features.
\begin{equation}
	\begin{aligned}
		x_{t1}=DWConv_{1 \times 1}(x_{F})	\\
		x_{t3}=DWConv_{3 \times 3}(x_{F})	\\
		x_{t5}=DWConv_{5 \times 5}(x_{F})
	\end{aligned}
\end{equation}
where $ DWConv(*) $ denotes the depth-wise convolution. Finally, all the obtained features are fused and restored to the original channel dimensions by the convolution with kernel size 1.
\begin{equation}
	\begin{aligned}
		x_{afem}=Conv_{1 \times 1}(ReLU(x_{t1}+x_{t3}+x_{t5}))
	\end{aligned}
\end{equation}
where $ ReLU $ \cite{hendrycks2016gaussian} represents a non-linear function. The Hash Layer transforms image features into hash codes, which contains one layer with a TANH non-linearity.

Since our AFEM module aims to increase feature diversity, the channel enhancement ratio needs to be determined. The generic method is to perform manual tuning via cross-validation, however, this costs a lot of computational resources. Group convolution has been successfully utilized to improve CNN architectures \cite{xie2017aggregated,zhang2017interleaved} where high-dimensional (low-dimensional) channels involve long-range (short-range) convolution given the fixed number of groups. Inspired by the results, we argue that the channel enhancement ratio should be proportional to the length of the hash code. In other words, there may exist a mapping relationship $ \phi $ between the channel increase ratio $ \lambda $ and the length of the hash code $ K $.
\begin{equation}
	\begin{aligned}
		\lambda = \phi(K)
	\end{aligned}
\end{equation}
The simplest mapping is a linear function, i.e.,  $\phi(K)=\mu * K + b$. Nevertheless, the relations characterized by linear function are too limited. Furthermore, it is well known that the channel dimension is usually set to a power of 2. By a similar principle, we introduce a possible solution by expanding the linear function into a nonlinear function, i.e.,
\begin{equation}
	\begin{aligned}
		\lambda = \phi(K)=2^{(\mu * K + b)}
	\end{aligned}
\end{equation}
Then, given the length of hash code $ K $, the channel enhancement ratio $ \lambda $ can be adaptively determined. In this work, we set $ \mu $ to $ \dfrac{1}{16} $ and $ b $ to $ 0 $ in all experiments, respectively. It is obvious from Eq. 16 that longer hash code correspond to larger channel dimensions for increasing greater diversity of features.

\begin{table*}[ht]
	\centering
	\caption{The corresponding results (MAP) on the three benchmark datasets.}
	\label{tab:commands}
	\begin{tabular}{c|cccccccccccc}
		\hline
		Datasets & \multicolumn{4}{c}{CIFAR-10@54000} & \multicolumn{4}{c}{NUS-WIDE@5000} & \multicolumn{4}{c}{IMAGENET@1000} \\
		\hline
		Methods & 16 bits & 32 bits & 48 bits & 64 bits & 16 bits & 32 bits & 48 bits & 64 bits & 16 bits & 32 bits & 48 bits & 64 bits \\
		\hline
		SH \cite{weiss2008spectral} & - & - & - & - & 0.4058 & 0.4209 & 0.4211 & 0.4104 & 0.2066 & 0.3280 & 0.3951 & 0.4191
		\\
		ITQ \cite{gong2012iterative}& - & - & - & - & 0.5086 & 0.5425 & 0.5580 & 0.5611 & 0.3255 & 0.4620 & 0.5170 & 0.5520\\
		KSH \cite{liu2012supervised}& - & - & - & - & 0.3561 & 0.3327 & 0.3124 & 0.3368 & 0.1599 & 0.2976 & 0.3422 & 0.3943\\
		BRE \cite{kulis2009learning}& - & - & - & - & 0.5027 & 0.5290 & 0.5475 & 0.5546 & 0.0628 & 0.2525 & 0.3300 & 0.3578\\
		\hline
		DSH \cite{liu2016deep}& 0.6145 & 0.6815 & 0.6828 & 0.6910 &0.6338 &0.6507 &0.6664 &0.6856 &0.4025 &0.4914 &0.5254 &0.5845 \\
		DHN \cite{zhu2016deep}& 0.6544 &0.6711 &0.6921 &0.6737 &0.6471 &0.6725 &0.6981 &0.7027 &0.4139 &0.4365 &0.4680 &0.5018 \\
		DPSH \cite{li2015feature}& 0.7230 &0.7470 &0.7550 &0.7750 & 0.7156 &0.7302 &0.7426 &0.7172 & 0.4531 &0.4836 &0.5020 &0.5330\\
		HashNet \cite{cao2017hashnet}& 0.7321 &0.7632 &0.7820 &0.7912 &0.6612 &0.6932 &0.7088 &0.7231 &0.4385 &0.6012 &0.6455 &0.6714 \\
		DCH \cite{cao2018deep}&0.7562 &0.7874 &0.7929 &0.7935 &0.7012 &0.7345 &0.7306 &0.7151 &0.4356 &0.5663 &0.5872 &0.5688 \\
		MMHH \cite{kang2019maximum}& 0.7956 &0.8087 &0.8152 &0.8178 & 0.7687 &0.7874 &0.7801 &0.7514 & - & - & - & - \\
		DPN \cite{fan2020deep}&0.8250 &0.8380 &0.8300 &0.8290 & - & - & - & - & 0.6840 &0.7400 &0.7560 &0.7610\\
		TransHash \cite{chen2022transhash}& 0.9075 &0.9108 &0.9141 &0.9166 &0.7263 &0.7393 &0.7532 &0.7488 &0.7852 &0.8733 &0.8932 &0.8921 \\
		HashFormer \cite{li2022hashformer}&0.9121 &0.9167 &0.9211 &0.9236 &0.7317 &0.7418 &0.7592 &0.7597 &0.7791 & 0.8962 &0.9007 &0.9010 \\
		MSViT-B \cite{li2023msvit}&0.8982 &0.9281 &0.9380 &0.9443 & - & - & - & - &0.7869 &0.8635 &0.8926 &0.9036 \\
		HybridHash \cite{he2024hybridhash} & 0.9367  & 0.9413   & 0.9468  & 0.9513 & 0.7785   & 0.7986    & 0.8068   & 0.8164 & 0.8028  & 0.8886   & \textbf{0.9094}  & 0.9110 \\
		MambaHash(ours) &\textbf{0.9438}  &\textbf{0.9483}   &\textbf{0.9500}  &\textbf{0.9530} &\textbf{0.7786}   &\textbf{0.7991}    &\textbf{0.8075}   &\textbf{0.8178} &\textbf{0.8133}  & \textbf{0.8974}   & 0.9087  &\textbf{0.9133} \\
		
		\hline
	\end{tabular}
\end{table*}

\subsection{Objective Function}
To preserve the similarity information of the pairwise images, we adopt negative log-likelihood \cite{li2015feature} for guiding the model to generate binary hash codes. Given training
images ($ \boldsymbol{h}_i, \boldsymbol{h}_j,  s_{ij} $). Here, $ \mathcal{S} = \left\{s_{ij}\right\}$ represents the similarity matrix where $ s_{ij}=1 $ if $ \boldsymbol{h}_i $ and $ \boldsymbol{h}_j $ are from the same class and $ s_{ij}=0 $ otherwise. we can get the following optimization problem:
\begin{equation}
	\begin{aligned}
		\min _{\mathcal{H}} \mathcal{J}_{1} & =-\log p(\mathcal{S} \mid \mathcal{H})=-\sum_{s_{i j} \in \mathcal{S}} \log p\left(s_{i j} \mid \boldsymbol{h}_i,\boldsymbol{h}_j\right) \\
		& =-\sum_{s_{i j} \in \mathcal{S}}\left(s_{i j} \theta_{i j}-\log \left(1+e^{\theta_{i j}}\right)\right)
	\end{aligned}
\end{equation}
where $ \mathcal{H} = \left\{\boldsymbol{h}_1, \ldots \boldsymbol{h}_N\right\} $, $ \theta_{i j}= \dfrac{1}{2}\boldsymbol{h}_i^\top\boldsymbol{h}_j $. Sicne the inner product $ \left(\boldsymbol{h}_i^\top\boldsymbol{h}_j\right) $ and the Hamming distance $ H\left(\boldsymbol{h}_{i}, \boldsymbol{h}_{j}\right) $ have a nice relationship, i.e., $ H\left(\boldsymbol{h}_{i}, \boldsymbol{h}_{j}\right)=\dfrac{1}{2}\left(K-\left(\boldsymbol{h}_i^\top\boldsymbol{h}_j\right)\right) $. It is evident to notice that the above optimization problem makes the Hamming distance between the similar image pairs as small as possible and the Hamming distance between the dissimilar image pairs as large as possible. To reduce the quantization error, we add a quantization error loss, then the entire optimization objective is:
\begin{equation}
	\begin{aligned}
		\min _{\mathcal{H}} \mathcal{J}_{2} = & -\sum_{s_{i j} \in \mathcal{S}}\left(s_{i j} \theta_{i j}-\log \left(1+e^{\theta_{i j}}\right)\right) \\
		& + \eta \sum_{i=1}^{n}\left\|\boldsymbol{h}_{i}-sign(\boldsymbol{h}_{i})\right\|_{2}^{2}
	\end{aligned}
\end{equation}
where $ \eta $ is the hyper-parameter and $ sign(*) $ denotes the sign function.

\section{Experiments}
\subsection{Datasets and Evaluation Protocols}
We have conducted experiments on three widely used datasets, which are CIFAR-10, NUS-WIDE and IMAGENET.

\textbf{CIFAR-10}: CIFAR-10 is a single-labeled dataset with a total of 60,000 images, of which 50,000 are used for training and 10,000 for testing. The dataset has 10 categories with 6,000 images in each category. We follow the same setup as the experiment in \cite{chen2022transhash}. 1000 images are used as the test set, 5000 images are randomly selected as the training set, and the remaining 54000 images are treated as the image database (retrieval set).

\textbf{NUS-WIDE}: NUS-WIDE is a multi-labeled dataset which is commonly used for large-scale image retrieval tasks. The dataset contains 269,648 images with 81 categories. We follow the experimental setup in \cite{chen2022transhash}. 5,000 images are utilized as the test set and 168,692 images are treated as the retrieval set after removing the images with noise. Subsequently 10,000 images were randomly selected from the retrieval set as the training set. 

\textbf{IMAGENET}: IMAGENET is the benchmark image dataset for the Large Scale Visual Recognition Challenge (ILSVRC 2015). Specifically, we follow the experimental setup in \cite{chen2022transhash} and randomly select 100 category images. These 100 categories have 128,503 training images, which are regarded as the retrieval set, and 5,000 test images are utilized as the query set. Finally, 130 images from each category are randomly selected as the training set.

We adopt the mean average precision (MAP) of different bits $ \left\{16, 32, 48, 64\right\} $ to evaluate the quality of the retrieved images. Concretely, we followed similar work \cite{chen2022transhash,cao2017hashnet} and the MAP results were calculated based on the top 54,000 returned samples from the CIFAR-10 dataset, 5,000 returned samples from the NUS-WIDE dataset, and 1,000 returned samples from the IMAGENET dataset.

\begin{table*}[ht]
	\centering
	\caption{The corresponding results (MAP) of different variants of MambaHash on the three benchmark datasets.}
	\label{tab:commands}
	\begin{tabular}{c|cccccccccccc}
		\hline
		Datasets & \multicolumn{4}{c}{CIFAR-10@54000} & \multicolumn{4}{c}{NUS-WIDE@5000} & \multicolumn{4}{c}{IMAGENET@1000} \\
		\hline
		Methods & 16 bits & 32 bits & 48 bits & 64 bits & 16 bits & 32 bits & 48 bits & 64 bits & 16 bits & 32 bits & 48 bits & 64 bits \\
		\hline
		MambaHash &\textbf{0.9438}  &\textbf{0.9483}   &\textbf{0.9500}  &\textbf{0.9530} &\textbf{0.7786}   &\textbf{0.7991}    &\textbf{0.8075}   &\textbf{0.8178} &\textbf{0.8133}  & \textbf{0.8974}   &\textbf{0.9087}  &\textbf{0.9133} \\
		MambaHash w/o C & 0.9228   & 0.9398   & 0.9408    & 0.9443   & 0.7666  & 0.7870  & 0.7977  & 0.8046  &  0.8031    & 0.8855  & 0.9047    & 0.9061    \\
		MambaHash w/o F & 0.9317  & 0.9362   & 0.9446  & 0.9442 & 0.7587   & 0.7832    & 0.7993   & 0.8058 & 0.8023  & 0.8847   & 0.9056  & 0.9069 \\
		MambaHash* &0.9177 &0.9192 &0.9245 &0.9333 &0.7536 &0.7811 &0.7925 &0.7972 &0.7915 & 0.8721 &0.9026 &0.9048 \\

		\hline
	\end{tabular}
\end{table*}

\subsection{Implementation Details}
All images are originally resized to 256 × 256. For the training images, we use standard image augmentation techniques, including random horizontal flipping and random cropping with the crop size of 224. To speed up the convergence of the model, we utilize RMSProp as the optimizer. For the experimental parameters, the batch size is $ 32 $, the learning rate is tuned in the range of $ \left[1 \times 10^{-5}, 2 \times 10^{-5}\right] $ and the weight decay parameter is set to $ 10^{-7} $. We obtained the hyperparameter a of MambaHash by cross-validation to be 0.01, 0.05, and 0.05 on the CIFAR-10,  NUS-WIDE and IMAGENET datasets, respectively. All experiments are conducted with one Tesla V100 GPU.

\subsection{Experimental Results and Analysis}
In this section, we compare the performance of our proposed MambaHash with state-of-the-art deep hashing methods. Concretely, the compared methods can be grouped into two categories: hand-crafted based hashing methods and deep learning based hashing methods. For the hand-crafted based hashing methods, we select the more typical methods SH \cite{weiss2008spectral}, ITQ \cite{gong2012iterative}, KSH \cite{liu2012supervised} and BRE \cite{kulis2009learning} for performance comparison. For deep learning based hashing methods, we further include DSH \cite{liu2016deep}, which is one of the first works for deep convolutional neural networks dealing with the image retrieval hashing problem. Furthermore, we incorporate other state-of-the-art deep hashing methods, including DHN \cite{zhu2016deep}, DPSH \cite{li2015feature}, HashNet \cite{cao2017hashnet}, DCH \cite{cao2018deep}, MMHH \cite{kang2019maximum}, DPN \cite{fan2020deep}, TransHash \cite{chen2022transhash}, HashFormer \cite{li2022hashformer}, MSViT \cite{li2023msvit} and HybridHash \cite{he2024hybridhash}. It is important to emphasize that all the results of non-deep-learning hashing methods and deep-learning hashing methods are derived from \cite{he2024hybridhash}.

Table 1 illustrates the MAP results of different hashing methods on the three benchmark datasets. It can be evident that for the hand-crafted based hashing method, the deep learning based hashing method has a significant performance improvement. The reason may be that hand-crafted visual descriptors are difficult to guarantee the similarity between image pairs, resulting in the generation of sub-optimal hash codes.  Although the deep learning based hashing methods achieve superior performance on the three benchmark datasets, our method still meets or even exceeds the state-of-the-art hashing methods. The reasons are mainly summarized in two points. First, the image scanning with Mamba in all four directions can effectively capture the feature information. The second is that the adaptive feature enhancement module indeed increases the feature diversity and enhances the visual representation of the model. It also illustrates that with well designed architecture, Mamba can fulfill the image retrieval task well. Moreover, our method also consistently outperforms with superior performance on NUSWIDE dataset for different hash bit lengths. This demonstrates that MambaHash is also suitable for multi-label image retrieval, where each image contains multiple labels.

\subsection{Model Settings}
The MambaHash accepts pairs of input images of size $ 224 \times 224 $. The whole model consists of four stages with stacked standard mamba blocks of numbers  ($ L_1 $) 3,  ($ L_2 $) 4,  ($ L_3 $) 16, and  ($ L_4 $) 3 and hidden feature dimensions of 64, 128, 348, and 512. For the kernel size $ k $ in the CIAM module, we set to 3, 3, 5, and 5 at four different hash bit levels with reference to the method in \cite{wang2020eca}.

\subsection{Ablation Study}
To further analyze the overall design of our proposed method, a detailed ablation study is performed to illustrate the effectiveness of each component. Specifically, we investigated two variants of MambaHash:
\begin{itemize}
	\item \textbf{MambaHash*}: A variant that contains only Mamba blocks without adding additional modules.
	\item \textbf{MambaHash w/o C}: A variant without adopting the Channel Interaction Attention Module (CIAM).
	\item \textbf{MambaHash w/o F}: A variant without adopting the Adaptive Feature Enhancement Module (AFEM).
\end{itemize}

\begin{table}[t]
	\caption{Comparison of MAP results for different backbone networks on three benchmark datasets.}
	\label{tab:freq}
	\begin{tabular}{c|c|ccccc}
		\hline
		Dataset & Bits & ResNet50 & ViT-L/32 & Hybrid & Mamba\\ 
		\hline
		\multirow{4}{*} {\makecell[c]{CIFAR-10\\@54000}}
		& 16  & 0.8315  & 0.8512 &  0.9367 & \textbf{0.9438}\\
		& 32  & 0.8465  & 0.8754 &  0.9413 & \textbf{0.9483}\\
		& 48  & 0.8543  & 0.8847 &  0.9468 &  \textbf{0.9500}\\
		& 64  & 0.8621  & 0.8969 &  0.9513 &  \textbf{0.9530}\\
		
		\hline
		\multirow{4}{*}{\makecell[c]{NUS-WIDE\\@5000}} 
		& 16  & 0.7147   &0.7152 &  0.7785 & \textbf{0.7786}\\
		& 32  & 0.7365    & 0.7411& 0.7986 & \textbf{0.7991}\\
		& 48  & 0.7487   & 0.7624&  0.8068 &  \textbf{0.8075}\\
		& 64  & 0.7566   & 0.7753&  0.8164 &  \textbf{0.8178}\\
		
		\hline
		\multirow{4}{*}{\makecell[c]{IMAGENET\\@1000}} 
		& 16  & 0.7384 &0.7477 &   0.8028 &  \textbf{0.8133}\\
		& 32  & 0.7452 &0.7781 &   0.8886 &  \textbf{0.8974}\\
		& 48  & 0.7868 & 0.8087& \textbf{0.9094} &  0.9087\\
		& 64  & 0.8112 & 0.8265&   0.9110 &  \textbf{0.9133}\\
		
		\hline
	\end{tabular}
\end{table}

\begin{table}[t]
	\caption{Comparison of the efficiency for different backbone networks on three benchmark datasets with 48-bit length hash codes.}
	\label{tab:freq}
	\begin{tabular}{c|cc}
		\hline
		backbone & \#param & FLOPs \\ 
		\hline
		ResNet50 & 23.66M & 4.14G \\
		ViT-L/32 & 292.93M & 14.76	G \\
		Hybrid & 55.57M & 14.17G \\
		Mamba  &  38.99M &  7.53G \\
		\hline
	\end{tabular}
\end{table}

Table 2 exhibits the MAP performance of MambaHash and its variants on the three datasets. As can be noticed from Table 2, we experience notable performance declines on all three datasets when the channel interaction attention module is removed (\textbf{MambaHash w/o C}). This demonstrates it is beneficial for the channels to communicate information with each other. We also see a performance decline when the adaptive feature enhancement module is removed (\textbf{MambaHash w/o F}). This indicates that the adaptive feature enhancement module indeed enhances the visual representation capability of the model. Furthermore, we have included the performance of model \textbf{MambaHash*} which removes the two modules as mentioned above. It can be observed that, in the first place, \textbf{MambaHash*} has good performance on its own, which illustrates that Mamba can indeed be applied to image retrieval tasks. In the second place, the two modules can simultaneously exist and work together to improve the performance of the model. 

We further investigate the mainstream backbone networks utilized by existing deep hashing methods, including convolutional neural networks (exemplified by ResNet50 \cite{he2016deep}), Transformer (exemplified by ViT-L/32 \cite{dosovitskiy2020image}), and hybrid networks (exemplified by HybirdHash \cite{he2024hybridhash}). The number of parameters, computational complexity and performance of MambaHash are compared with these backbone networks, as shown in Tables 3 and 4. As can be viewed from the table, our proposed MambaHash has almost optimal performance on three datasets. The number of parameters and computational complexity are slightly higher than ResNet50. While ResNet50 has less number of parameters and lower computational cost, the performance is the lowest of all backbone networks. Therefore, MambaHash has both efficiency and performance.

\section{Conclusion}
To explore the suitability of Mamba for large-scale image retrieval tasks, we propose a visual state space deep hashing model (MambaHash). Specifically, MambaHash adopts stage-wise structural design and utilizes Mamba to scan images along four groups of channels in multiple directions for efficiently modeling local and global information. Next, we propose a channel interaction attention module to enhance cross channel communication. Finally, we elaborately design an adaptive feature enhancement module to increase feature diversity and enhance the visual representation capability of the model. We have conducted extensive experiments on three benchmark datasets, and the experimental results demonstrate that compared with the existing state-of-the-art hashing methods, MambaHash has both efficiency and performance to effectively accomplish image retrieval tasks.

\begin{acks}
This paper is supported by the National Natural Science Foundation of China under Grant 62466040,  and the Natural Science Foundation of Inner Mongolia Autonomous Region under Grant 2024MS06029.
\end{acks}

\bibliographystyle{ACM-Reference-Format}
\balance
\bibliography{sample-base}

\end{document}